\documentclass[article]{elsarticle}

\usepackage{hyperref}

\usepackage{geometry}

\usepackage{times}
\usepackage{helvet}
\usepackage{courier}
\usepackage{url}
\usepackage{latexsym}
\usepackage{graphicx}
\usepackage{color}
\usepackage{CJK}
\usepackage{fancyhdr}
\usepackage{multirow}
\usepackage{amsfonts}
\usepackage{amsmath}
\usepackage{booktabs}
\usepackage{algorithm}
\usepackage{algpseudocode}

\journal{Journal of \LaTeX\ Templates}

\begin{document}

\begin{frontmatter}

\title{Commonsense Knowledge Enhanced Embeddings for Solving Pronoun Disambiguation Problems in Winograd Schema Challenge}


\author[addr1]{Quan Liu}
\ead{quanliu@mail.ustc.edu.cn}

\author[addr2]{Hui Jiang}
\ead{hj@cse.yorku.ca}

\author[addr1]{Zhen-Hua Ling}
\ead{zhling@ustc.edu.cn}

\author[addr3]{Xiaodan Zhu}
\ead{zhu2048@gmail.com}

\author[addr4]{Si Wei}
\ead{siwei@iflytek.com}

\author[addr1,addr4]{Yu Hu}
\ead{yuhu@iflytek.com}

\address[addr1]{National Engineering Laboratory for Speech and Language Information Processing\\
University of Science and Technology of China, Hefei, China}
\address[addr2]{Department of Electrical Engineering and Computer Science, York University, Canada}
\address[addr3]{National Research Council Canada, Ottawa, Canada}
\address[addr4]{iFLYTEK Research, Hefei, China}

\begin{abstract}
In this paper, we propose commonsense knowledge enhanced embeddings (KEE) for solving the Pronoun Disambiguation Problems (PDP).
The PDP task we investigate in this paper is a complex coreference resolution task which requires the utilization of commonsense knowledge.
This task is a standard first round test set in the 2016 Winograd Schema Challenge\footnote{\url{http://commonsensereasoning.org/disambiguation.html}}.
In this task, traditional linguistic features that are useful for coreference resolution, e.g. context and gender information, are no longer effective anymore.
Therefore, the KEE models are proposed to provide a general framework to make use of commonsense knowledge for solving the PDP problems.
Since the PDP task doesn't have training data, the KEE models would be used during the unsupervised feature extraction process.
To evaluate the effectiveness of the KEE models, we propose to incorporate various commonsense knowledge bases, including ConceptNet, WordNet, and CauseCom, into the KEE training process.
We achieved the best performance by applying the proposed methods to the 2016 Winograd Schema Challenge.
In addition, experiments conducted on the standard PDP task indicate that, the proposed KEE models could solve the PDP problems by achieving 66.7\% accuracy, which is a new state-of-the-art performance.
\end{abstract}

\begin{keyword}
Winograd Schema Challenge (WSC) \sep Knowledge Enhanced Embeddings \sep Pronoun Disambiguation Problems (PDP) \sep Commonsense Knowledge \sep Knowledge Representation
\end{keyword}

\end{frontmatter}

\section{Introduction}

In recent years, many artificial intelligence (AI) challenges and competitions have been proposed for evaluating the intelligence of computers \cite{levesque2011winograd,weston2015towards,clark2015elementary}.
Among those challenges, the Winograd Schema Challenge (WSC) has been proposed as an alternative to the Turing Test \cite{levesque2011winograd}.
Turing first introduced the notion of testing a computer system's intelligence by assessing whether it could make a human judge think that she was conversing with a human rather than a computer \cite{turing1950computing}.
However, some recent efforts have leveraged surface conversation tricks to fool humans in believing they are speaking to another human being but not a computer \cite{veseloveugene,warwick2014good}.
To address this issue, the Winograd Schema Challenge is claimed to be a more reasonable task that provides an effective framework for testing a system's ability to understand natural language and use commonsense knowledge.
A Winograd schema (WS) question is a pair of sentences that differ only in one or two words which results in a different resolution of coreference.
A typical example \cite{levesque2011winograd} is the following:
\begin{description}
	\item[] Example WS:\\
	\emph{The customer walked into the bank and stabbed one of the tellers. He was immediately taken to the emergency room.}
	\\ \underline{Who} was taken to the emergency room? The customer/ the teller.
\end{description}
The correct answer is \textit{the teller}. We know this because of all the commonsense knowledge that we have
about stabbings, injuries, and how they are treated. We know that if someone is stabbed, he is very likely to
be seriously wounded, and that if someone is seriously wounded, he needs medical attention.
We also understand that people with acute and serious injuries are often treated at emergency rooms.
Moreover, there is no indication in the text that the customer has been injured, and therefore no apparent
reason for him to be taken to the emergency room. We reason with much of this information when we
determine that the referent of ``who" in the second sentence above is the teller rather than the customer.

The PDP task is designed as the first round test set of the 2016 Winograd Schema Challenge \cite{morgenstern2016planning}.
Because creating Winograd schemas is difficult, requiring creativity and inspiration, and too burdensome to do on a yearly or biennial basis, the Winograd Schema Challenge does not target directly at WS but focus first on a PDP task.
One of the original purposes of Winograd schemas, i.e. the correct answer be dependent on commonsense knowledge, is retained during the construction process of PDP problems.
In addition, strong preference will be given to PDP that do not rely on selectional restriction or on syntactical characteristics of corpora,
and which are of roughly the same complexity as Winograd schemas.
A typical PDP example is ``Mrs. March gave the mother tea and gruel, while she dressed the little baby as tenderly as if it had been her own."
One way to reason that \underline{she} in \textit{she dressed} refers to Mrs. March and not the mother, is to realize that the phrase ``as if it were her own" implies that it (the baby) is not actually her own; that is, \underline{she} is not the mother and must, by process of elimination, be Mrs. March.
Similar to the Winograd schemas, a substantial amount of commonsense knowledge appears to be necessary to disambiguate pronouns.

The PDP task doesn't have direct training data. By applying state-of-the-art coreference resolvers to answer the PDP problems, we find the
performances are very poor (close to a random guess).
\textbf{Therefore, employing commonsense knowledge is a urgent request, which is actually a goal that WS and PDP problems are designed for.}
This paper proposes to utilize commonsense knowledge for extracting representative features to make decisions.
We call the proposed model commonsense knowledge enhanced embeddings (KEE).
In the KEE framework, the commonsense knowledge would be quantized as semantic constraints to guide the semantic word embedding training process.
Based on it, the KEE models could not only effectively learn useful semantic information from large text corpora, but also encode some basic commonsense knowledge.
After training KEE models on a large Wikipedia corpus under the supervision of various commonsense knowledge bases, including ConceptNet\cite{liu2004conceptnet}, WordNet, and CauseCom, this paper proposes to represent all the pronouns and candidate mentions in the PDP problems by composing their contexts based on the pre-trained KEE models.
Finally, the process to answer PDP problems could be fulfilled by directly calculating the semantic similarities between the representation vectors of the pronoun under concern and all candidate mentions.
The candidate with largest semantic similarity with respect to the pronoun will be predicted as the answer.

To further improve the system performances of solving PDP problems, we propose to train a mention pair classifier with neural networks using the features extracted from KEE models.
The \textit{mismatched} training data used in the paper is the widely used OntoNote dataset.
Different from the baseline system that makes decisions based on semantic similarities, this \textit{model training} method could help us to further utilize the existing resources in the coreference resolution community.
We conduct experiments on the official PDP datasets of the 2016 Winograd Schema Challenge \cite{morgenstern2016planning}.
Experimental results indicate that the proposed KEE models achieve the state-of-the-art performance.
The main contributions of this paper include:
\begin{itemize}
	\item This is the first work that investigates the effectiveness of commonsense knowledge bases when solving the novel PDP task proposed in the Winograd Schema Challenge.
	\item Since the PDP task doesn't have training data, we propose a flexible strategy to employ commonsense knowledge in an unsupervised feature extraction process. And the experimental results do prove the effectiveness of this method.
\end{itemize}

The remainder of the paper will start with introducing the main motivation.
After that, we introduce the main methods proposed to solve the complex pronoun disambiguation problems.
In this paper, we propose a general KEE framework to employ commonsense knowledge bases for extracting representative features.
Based on KEE, we further propose two methods to finally answer the PDP problems.
We then present all the experiments, including setup, datasets, and results, before we conclude this paper.

\section{The Research Motivation}
The motivation of this paper comes from the wide attentions to the Pronoun Disambiguation Problem (PDP), a complex coreference resolution task proposed in the 2016 Winograd Schema Challenge \cite{morgenstern2016planning}.
Similar to the Winograd Schema (WS) problems, solving PDP problems also requires commonsense reasoning.
In this work, we propose to investigate the effectivenesses of various commonsense knowledge bases (KBs) for tackling PDP problems in a general framework.
To make clear the main motivation of this work, we will first describe the problems we aim to solve.
After that, detailed descriptions of an overall framework would be presented in the coming section.
\subsection{Pronoun Disambiguation Problems (PDP)}
In this year, the pronoun disambiguation problems (PDP) are designed as the first round test set for the Winograd Schema Challenge\cite{morgenstern2016planning}.
The PDP problems are taken directly or modified
from examples found in literature, biographies, autobiographies, essays, news analyses, and news stories; or
have been constructed by the organizers of the competition.
According to the descriptions of paper \cite{morgenstern2016planning}, here are some typical PDP examples:
\begin{description}
	\item[] Example PDP 1:\\
	\emph{Mrs. March gave the mother tea and gruel, while she dressed the little baby as tenderly as if it had been her own.}\\
	\underline{She} dressed: Mrs. March / the mother\\
	As if \underline{it} had been: tea / gruel / baby
	\item[] Example PDP 2:\\
	\emph{Tom handed over the blueprints he had grabbed and, while his companion spread them out on his knee, walked toward
the yard.}\\
	\underline{His} knee: Tom / companion
	\item[] Example PDP 3:\\
	\emph{One chilly May evening the English tutor invited Marjorie and myself into her room.}\\
	\underline{Her} room: the English tutor / Marjorie
	\item[] Example PDP 4:\\
	\emph{Mariano fell with a crash and lay stunned on the ground. Castello instantly kneeled by his side and raised his head.}\\
	\underline{His} head: Mariano / Castello
\end{description}

A PDP can be taken directly from text (Example PDP 3 is taken from Vera Brittain’s autobiography
\textit{Testament of Youth}) or may be modified (Examples PDP 1, 2, and 4 are modified slightly from the
novels \textit{Little Women}, \textit{Tom Swift} and \textit{His Airship}, and \textit{The Pirate City: An Algerine Tale}).
Meanwhile, a pronoun disambiguation problem may consist of more than one sentence, as in Example PDP 4.
In practice, we will rarely use PDPs that contain more than three sentences.
Moreover, there may be multiple pronouns and therefore multiple ambiguities in a sentence, as in Example
PDP 1.

As in Winograd schemas, a substantial amount of commonsense knowledge appears to be necessary to
disambiguate pronouns in PDP.
For example, one way to reason that she in ``\textit{\underline{she} dressed}" (Example PDP 1) refers to
Mrs. March and not the mother, is to realize that the phrase ``\textit{as if it were her own}" implies that it (the baby)
is not actually her own; that is, \underline{she} is not the mother and must, by process of elimination, be Mrs. March.
Similarly one way to understand that the English tutor is the correct referent of \underline{her} in Example PDP 3 is
through one's knowledge of the way invitations work: X typically invites Y into X's domain, and not into
Z's domain.
Especially, X does not invite Y into Y's domain.
Similar knowledge of etiquette come into play
in Example PDP2: one way to understand that the referent of \underline{his} is Tom is through the knowledge that X typically spreads documents out over X's own knee, and not Y's knee.
(Other knowledge that comes
into play is the fact that a person doesn't have a lap while he is walking, and the structure of the sentence
entails that Tom is the individual who walks to the yard.)

A difference between PDP and WS problems is that, the number of candidate noun phrases in each PDP problem would not always be two, but can be three, four, or even more. Therefore, the random-guess accuracy in the PDP problems will be lower than 50\% while the accuracy of a random-guess for WS is 50\%.

\subsection{Motivation}
When we starting this work on the PDP task, we used state-of-the-art coreference resolution toolkits to solve the PDP problems. As expected, the performances of those toolkits on the PDP task are very poor, which are almost close to the random guess performance (45\% accuracy on the official PDP test set).
The results indicates that traditional linguistic features that are useful for standard coreference resolution are no longer useful for the PDP task.
This motivate us to investigate what are the major difficulties entailed in the PDP task.
This paper considers that \textbf{the difficulties of solving the complex PDP problems} mainly include:
\begin{itemize} 
\item \textbf{Reason 1: The lack of training data} \ Similar to the Winograd Schema problems, the PDP task doesn't have any training data. In the viewpoint of machine learning, we can do nothing due to the lack of training data.
\item \textbf{Reason 2: The requirement of commonsense reasoning} \ Solving PDP problems is not easy since it requires reasoning over commonsense knowledge. In the AI community, the lack of commonsense knowledge and the weakness of commonsense reasoning programs would make the task investigated in this paper extremely challenging \cite{davis2015commonsense}.
\end{itemize}
\begin{figure}[htb]
  \centering
  \includegraphics[width=13.5cm]{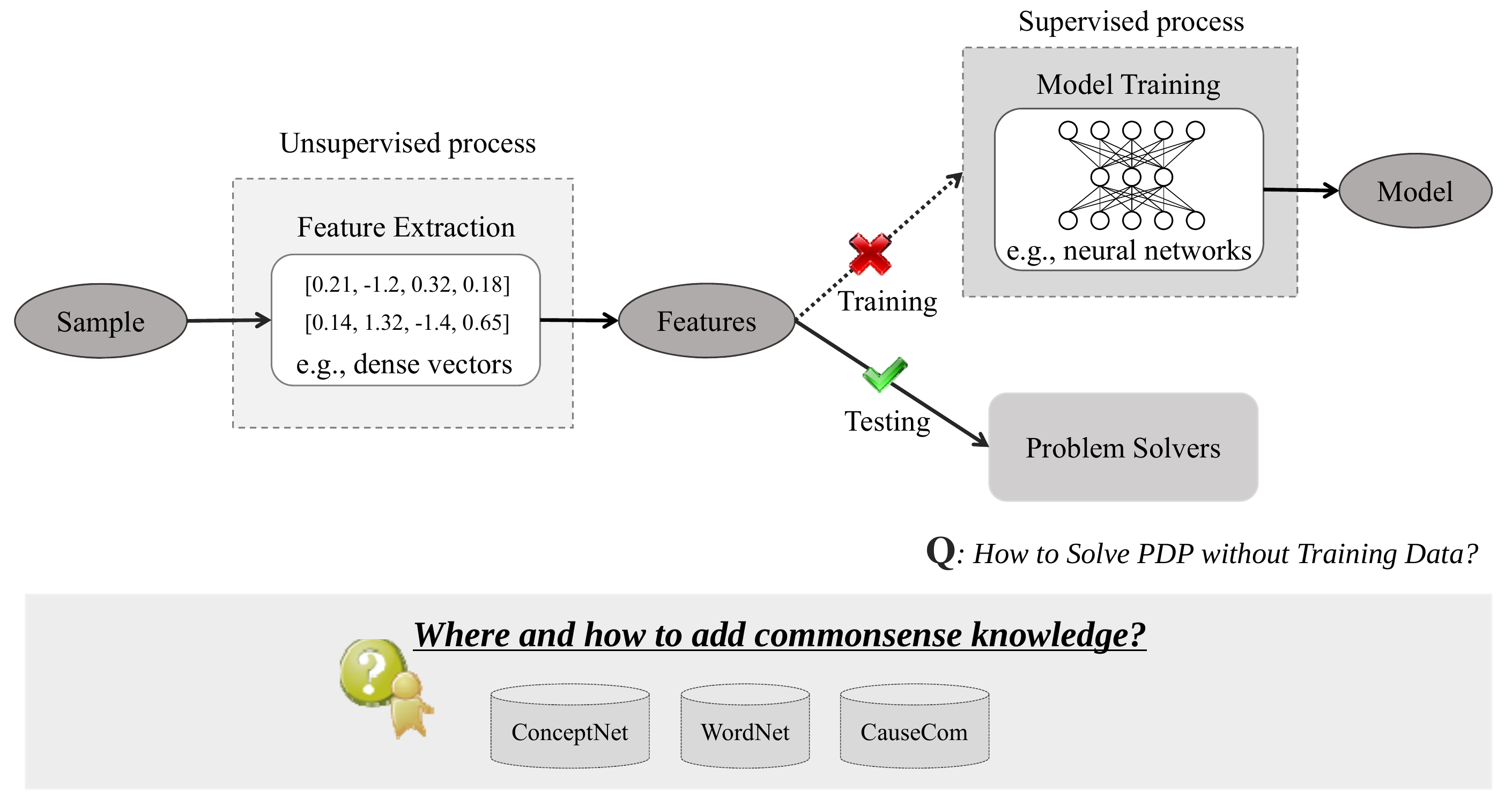}\\
  \caption{The main motivation of this paper. We aim to answer the two questions: 1) Where to add commonsense knowledge? 2) {How to Solve PDP without Training Data?} In the typical machine learning paradigm, to solve the PDP problems, we will first need to extract useful features from the input PDP examples. This process is a typical unsupervised process. After that, since we do not have any training data in the PDP task, we cannot train a straightforward model by applying typical machine learning algorithms, e.g. neural networks. Therefore, we can only choose to \textit{design standalone problem solvers}. In addition, since the process to address PDP problems requires commonsense knowledge, \textit{where to add commonsense knowledge} becomes a key issue as well.
  }
  \label{fig:motivation}
\end{figure}

This work is motivated by trying to solve these difficulties in the PDP problems. Figure \ref{fig:motivation} gives a straightforward illustration to our motivation.
As shown in Figure \ref{fig:motivation}, we put the process of solving PDP problems in the machine learning paradigm.
Given a PDP sample, the first thing we need to do is feature extraction, i.e. to extract representative features. Obviously, this process is an unsupervised process which do not rely on labelled training data.
Based on the extracted features, we used to have two choices, i.e., training models and conducting testing process directly.
However, in the PDP task, since we cannot obtain any training data, we have no way to train a model for resolving the PDP problems.
This is shown in the Figure \ref{fig:motivation} by a \textit{delete} symbol.
Due to this reason, we can only choose to solve the PDP problems by designing problem solvers, just as indicated in the Figure \ref{fig:motivation} by a \textit{checked} symbol.
Based on those analysis, a straightforward question emerges as \textit{How to Solve PDP without Training Data?}.
Moreover, since we need commonsense knowledge to answer PDP problems, a key question comes: \textit{Where and how to add commonsense knowledge?}
The candidate solutions to these two questions could be described in Table \ref{tab:motivation}.
\begin{table}[htb]
\centering
\begin{tabular}{l|c|c}
  \hline
  The question & Solutions & Checked\\\hline
  \textit{Where and how to add commonsense knowledge?} & \textit{feature extraction process} & $\surd$ \\
  \hline
  \multirow{3}{*}{\textit{How to solve PDP without training data?}} & \textit{unsupervised feature similarity method} & $\surd$ \\
  & \textit{standalone model training method} & $\surd$ \\
  & \textit{traditional logic reasoning method} & $\times$ \\
  \hline
\end{tabular}
\caption{The candidate solutions to the key questions investigated in this work. The rightmost column indicates the solution we selected to use in this work. }
\label{tab:motivation}
\end{table}

The answer to question ``\textit{where to add commonsense knowledge?}" is to employ commonsense knowledge during the unsupervised feature extraction process.
In this paper, since we find that traditional features that are useful for standard coreference resolution are not useful for the PDP task, \textbf{we decide to simply extract the context feature} for each pronoun and candidate mentions in the PDP problem.
By using context features, we could utilize the context distributional information (this information could be learned from large number of text corpora by using the popular word embedding apporach) that may be useful for solving PDP problems.
More importantly, since we need to train semantic word embeddings to extract context features, we can incorporate various commonsense knowledge bases into the word embedding learning process.
Therefore, \textbf{this paper proposes commonsense knowledge enhanced embeddings (KEE) to address the complex PDP problems} in \textit{a general distributed representation learning framework}.
By training KEE models in an unsupervised way, we can learn word embeddings that encode the commonsense knowledge.
After that, extracting features for each pronoun and candidate mentions in the PDP problems could be implemented by directly composing their context word vectors.

On the other hand, finding answers to question ``\textit{How to solve PDP without training data?}" remains to be a research problem.
A straightforward method is to use \underline{traditional logic reasoning method}, i.e., solving PDP problems using logic reasoning method. It relies on semantic parsing (parsing natural language into logic forms \cite{poon2009unsupervised,wang2015building}) and logic reasoning, which is very complex.
In the 2016 Winograd Schema Challenge, the best logic reasoning system performs just 48\% on the official PDP test set \cite{WSC2016}.
In this paper, we propose to use two simple and reasonable methods:
1) \underline{Unsupervised feature similarity method}. After extracting useful features for each pronoun and candidate mentions, the resolution process could be realized by calculating the similarities between each pronoun and all the corresponding candidate mentions. The mention with the largest similarity value to the pronoun (to be resolved) would be treated as the final answer;
2) \underline{Standalone model training method}. Mismatched training data to train standalone models, and answer PDP problems.
A typical process would be using existing coreference resolution databases, e.g. OntoNote to train a coreference resolver. Just as emphasized in this paper, state-of-the-art coreference resolvers are not effective when solving PDP problems.
That means, the key point of this method is to find or construct useful training data, which is out of scope of this paper.
In this paper, in order to investigate the effectiveness of commonsense knowledge enhanced embeddings for feature extraction, we will use both the \textit{unsupervised feature similarity method} and \textit{standalone model training method} to solve the PDP problems.
\begin{figure}[htb]
  \centering
  \includegraphics[width=12.5cm]{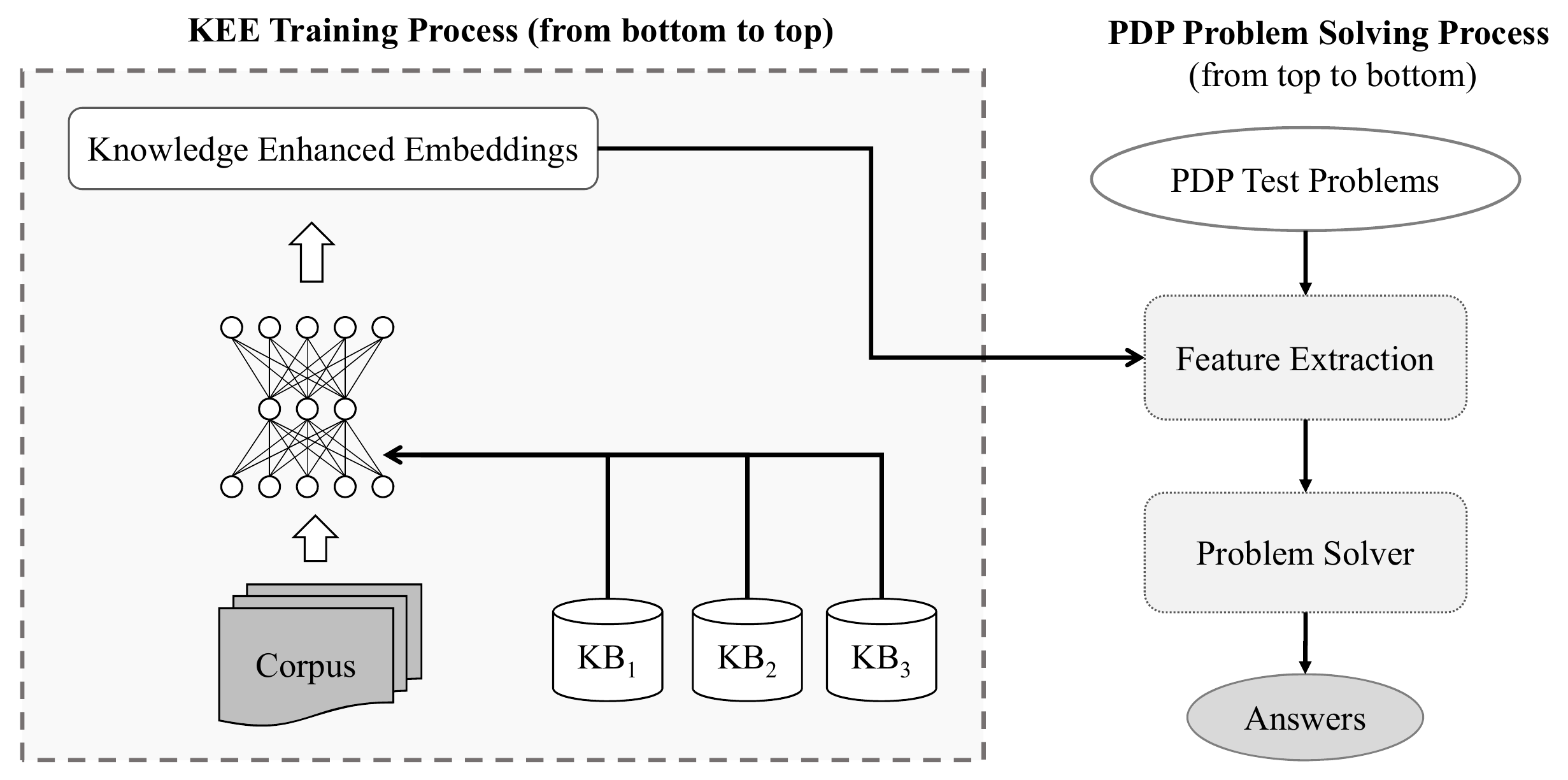}\\
  \caption{The overall system framework for solving PDP problems. The knowledge enhanced embeddings would play a key role in this work. In detail, the KEE model trained by combining text and various commonsense knowledge would be used during the feature extraction process. After feature extraction, the PDP test problems could be solved by employing problem solver. In this paper, we use the \textit{unsupervised feature similarity method} as our baseline solver.}
  \label{fig:kee-focus}
\end{figure}

\section{Commonsense Knowledge Enhanced Embeddings}
Based on the motivation, in this section, we introduce the Commonsense Knowledge Enhanced Embeddings (KEE) model in detail.
Before introducing the KEE model, we give an overall descriptions to the commonsense knowledge bases that are widely used in both the AI and NLP community.

\subsection{The Commonsense Knowledge Bases}
There have been open commonsense knowledge bases in the artificial intelligence community, e.g. ConcepNet \cite{liu2004conceptnet}, Cyc \cite{lenat1995cyc,mueller1998natural}and WordNet \cite{miller1995wordnet}.
Cyc is an artificial intelligence project that attempts to assemble a comprehensive ontology and knowledge base of everyday common sense knowledge, with the goal to enable AI applications to perform human-like reasoning.
Typical pieces of knowledge represented in the Cyc database are ``every tree is a plant" and ``plants die eventually".
However, since the commonsense knowledge in Cyc are represented by formal language, it would be difficult for us to use in our system.
Therefore, in this work, we decide to use three commmonsense knowledge bases, including ConceptNet \cite{liu2004conceptnet}, WordNet \cite{miller1995wordnet} and CauseCom \cite{liu2016probabilistic}.
\begin{figure}[htb]
  \begin{minipage}[b]{5cm}
    \centering
    \includegraphics[width=1\textwidth]{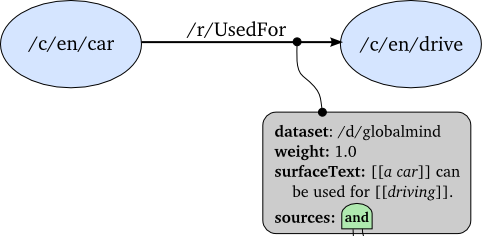}
    \caption{ConceptNet \cite{liu2004conceptnet}}
    \label{fig:cnet}
  \end{minipage}%
  \begin{minipage}[b]{4.5cm} 
    \centering
    \includegraphics[width=1\textwidth]{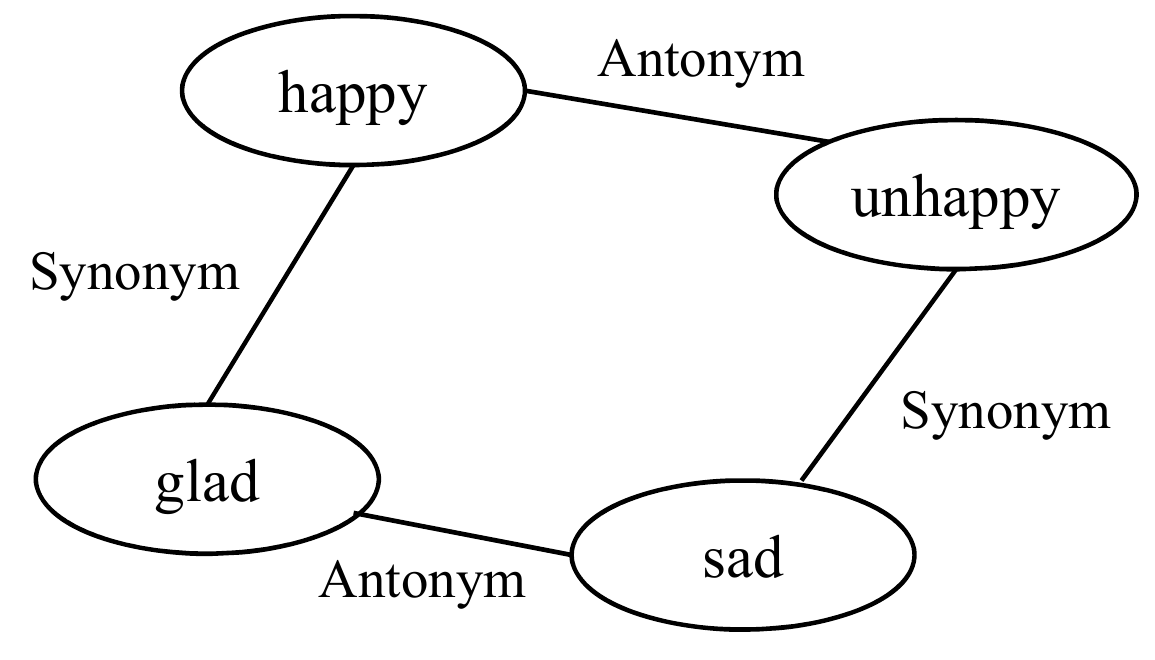}
    \caption{WordNet \cite{miller1995wordnet}}
    \label{fig:wnet}
  \end{minipage}
  \begin{minipage}[b]{5.75cm}
    \centering
    \includegraphics[width=1\textwidth]{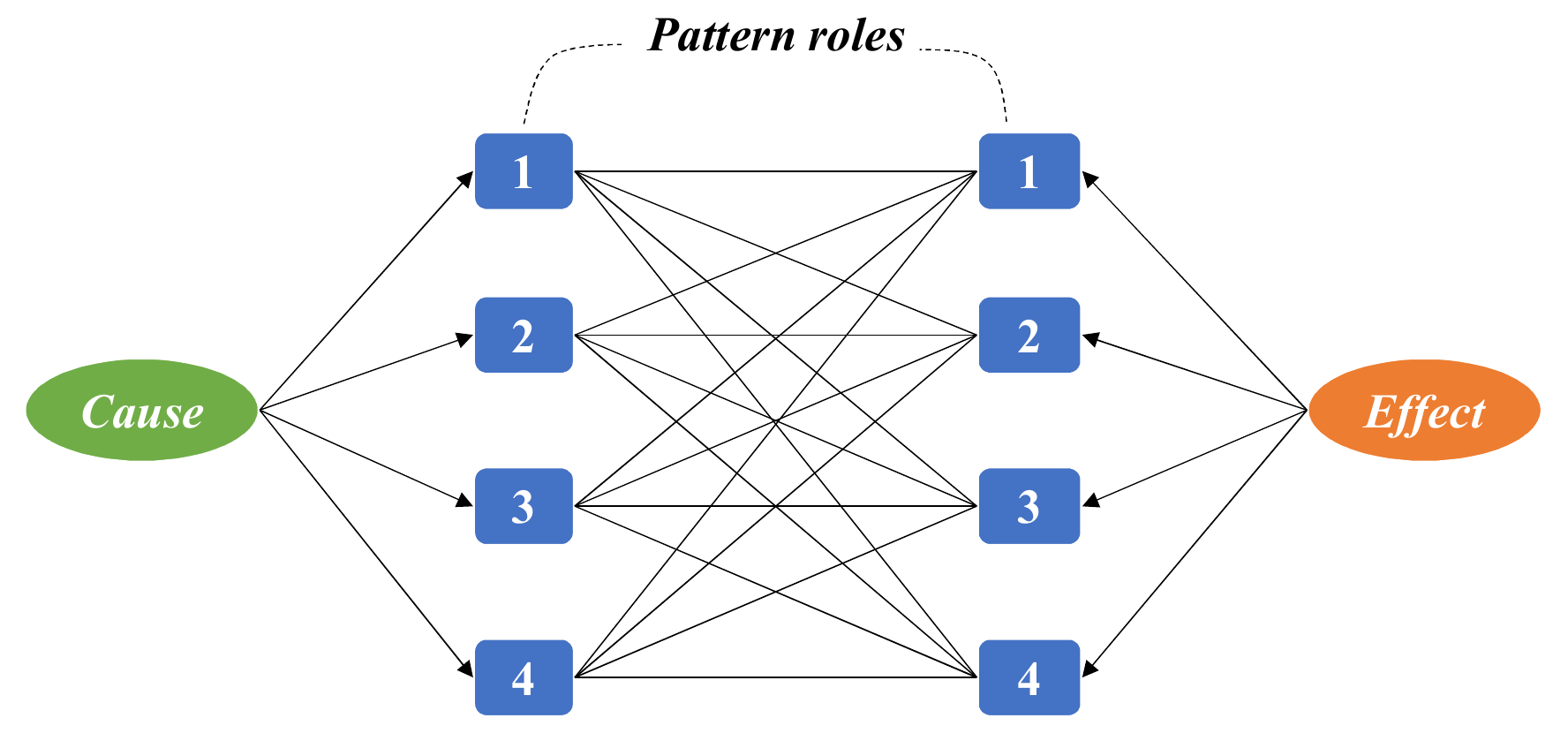}
    \caption{CauseCom \cite{liu2016probabilistic}.}
    \label{fig:causecom}
  \end{minipage}
\end{figure}

\textbf{ConceptNet} is a semantic network containing lots of things computers should know about the world, especially when understanding text written by people \cite{liu2004conceptnet}.
It is built from \textit{nodes} representing words or short phrases of natural language, and labeled \textit{relationships} between them.
For example, the triple (\textit{learn}, \textit{MotivatedByGoal}, \textit{knowledge}) indicates that ``we would learn because we want knowledge".
Moreover, as shown in Figure \ref{fig:cnet}, we can find the commonsense knowledge ``car is used for drive".
Those triples defined in ConceptNet would be used in the proposed KEE framework.

\textbf{WordNet} is a lexical database for the English language \cite{miller1995wordnet}. It groups English words into sets of synonyms called synsets, provides short definitions and usage examples, and records a number of relations among these synonym sets or their members. WordNet can thus be seen as a combination of dictionary and thesaurus.
WordNet includes the lexical categories nouns, verbs, adjectives and adverbs but ignores prepositions, determiners and other function words.
In the coming section, we will introduce how we extract useful knowledge from WordNet.

\textbf{CauseCom} is a set of cause-effect pairs automatically collected from large text corpus \cite{liu2016probabilistic}.
We call it \textit{CauseCom} for indicating this knowledge base is a \textbf{Cause} effect pair set of \textbf{Com}mon words.
Actually, the CauseCom KB is proposed to avoid data sparseness problem since the vocabulary of the KB covers common words (not phrases) in daily life, e.g., common verbs, adjectives, etc.
Figure \ref{fig:causecom} shows the typical formula of the corresponding KB.
As shown in Figure \ref{fig:causecom}, there are four pattern roles for both the \textit{cause} and \textit{effect} phrases.
The four roles include (active,positive), (active,negative), (passive,positive) and (passive,negative).
Typical example from this KB is: \textit{(win $\rightarrow$ happy)} which indicates \textit{win} would cause \textit{happy}.


\subsection{Knowledge Enhanced Embeddings (KEE)}
As described in the motivation section, this work proposes to incorporate commonsense knowledge into the word embedding process for learning knowledge enhanced embeddings.
In the natural language processing community, there are some works that have been proposed to employ semantic knowledge to improve the learning of word embeddings \cite{yu2014improving,faruqui2014retrofitting,bian2014knowledge,xu2014rc,liu2015learning}.
In this work, we propose to use a method similar to the one (named SWE) proposed in \cite{liu2015learning}. This method has been applied to many tasks, including word similarity evaluation, sentence completion, and semantic representation in brain activity \cite{liu2015learning,ruan2016emnlp}.
In the coming sections, we would first introduce the way we collect knowledge constraints from commonsense KBs.
After that, we introduce the main framework to learn knowledge enhanced embeddings.
\subsubsection{Knowledge constraints collected from commonsense KBs}
\label{sec:rules}
The key work of word representation learning is to learn the semantic similarities between all the words in vocabulary. In the KEE framework, the knowledge constraints are formulized as \textbf{semantic similarity inequalities} between two word pairs. Here we introduce the main methods we design to generate knowledge constraints from the three commonsense KBs used in this paper.
\begin{enumerate}[(1)]
\item \textbf{ConcepNet} \\
To make clear how to generate knowledge constraints for training KEE models, we need to know the input and output.
From ConceptNet, the input for generating knowledge constraints is a typical commonsense knowledge triple, i.e., $(w_{h}, r, w_{t})$ where $w_{h}$, $r$, and $w_{t}$ stands for head concept, relation, and tail concept respectively.
In this work, we design the following rules to generate knowledge constraints: Similarities between linked concepts should be larger than the similarities between unlinked concepts.
A typical inequality generated from triple $(w_{h}, r, w_{t})$ would be:
\begin{equation}\label{eq:ineq-cnet}
  (w_{h}, r, w_{t}) \Rightarrow \textrm{sim}(w_{h}, w_{t}) > \textrm{sim}(w_{h}, w_{k}), w_{k} \in V\ \mathrm{and}\ w_{k}\ \mathrm{is\ not\ linked\ with}\ w_{h}
\end{equation}
\item \textbf{WordNet}\\
In this work, we use the method proposed in \cite{liu2015learning} to generate semantic inequalities from WordNet:
1) Similarities between a word and its synonymous words are larger than similarities between the word and its antonymous words. A typical example is $\mathrm{similarity}(happy, glad) > \mathrm{similarity}(happy, sad)$.
2) Similarities of words that belong to the same semantic category would be larger than similarities of words that belong to different categories.
3) Similarities between words that have shorter distances in a semantic hierarchy should be larger than similarities of words that have longer distances.

\item \textbf{CauseCom}\\
Similar to the method designed for ConcepNet, given a cause-effect pair $(w_{i}, w_{j})$, we generate knowledge constraints by randomly sampling irrelevant words.
\begin{equation}\label{eq:ineq}
  (w_{i}, w_{j}) \Rightarrow \textrm{sim}(w_{i}, w_{j}) > \textrm{sim}(w_{i}, w_{k}), w_{k} \in V\ \mathrm{and}\ w_{k}\ \mathrm{is\ not\ the\ effect \ of}\ w_{i}
\end{equation}

The idea to generate such inequality is similar to the physical meaning of lexical entailment \cite{geffet2005distributional,turney2015experiments}.
Note that the commonsense knowledge base used in this paper covers all the common verbs and adjectives, and the knowledge constraints would not influence the learning for the remaining words in the whole large vocabulary.
The size of the common verbs and adjectives is 7500. Typical words are ``\textit{eat}", ``\textit{thank}", ``\textit{famous}", etc.
This is important because the used verbs and adjectives play a central role in commonsense reasoning.
Currently, incorporating more knowledge of other types of words, e.g., nouns, adverbs and prepositions is beyond the scope of this work.
\end{enumerate}

\subsubsection{The knowledge enhanced embedding framework}
\begin{figure}[htb]
  \centering
  \includegraphics[width=9cm]{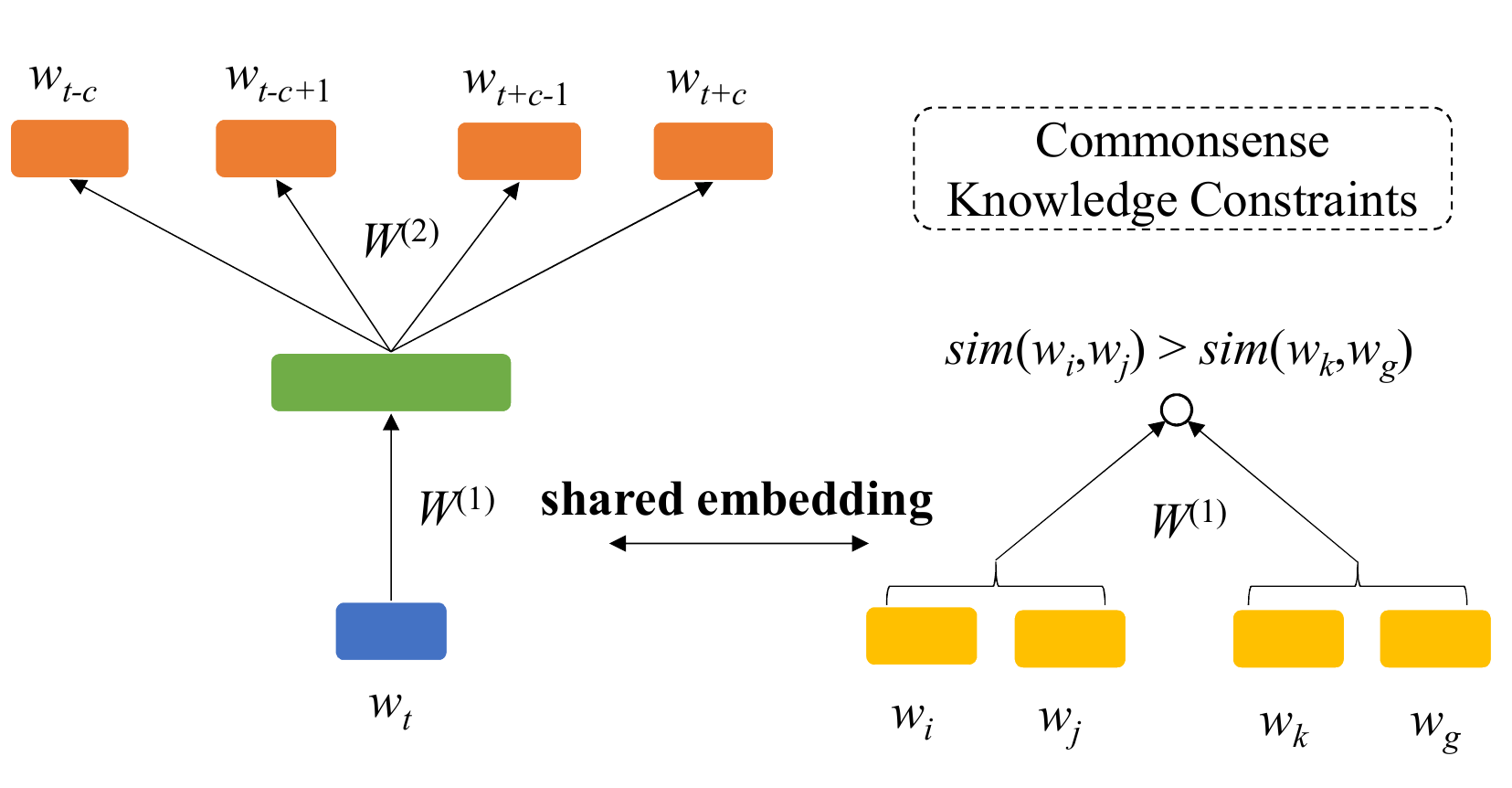}\\
  \caption{The framework for knowledge enhanced embeddings (KEE). After extracting semantic similarity inequalities from various commmonsense knowledge bases, we apply them to the word embedding training process. These inequalities works as knowledge constraints and are expected to be useful for helping us to learn better word representations.}
  \label{fig:swe-cssk}
\end{figure}

The main framework for learning knowledge enhanced embedding is shown in Figure \ref{fig:swe-cssk}.
The left part in this framework is the typical skip-gram model, which learns continuous word vectors from text corpora based on the aforementioned distributional hypothesis \cite{mikolov2013efficient}.
Each word in vocabulary (size of $V$) is mapped to a continuous embedding space by looking up an embedding matrix ${\bf W}^{(1)}$. And ${\bf W}^{(1)}$ is learned by maximizing the prediction probability, calculated by another prediction matrix ${\bf W}^{(2)}$, of its neighbouring words within a context window.

Given a sequence of training data, denoted as $w_{1}, w_{2}, w_{3}, ..., w_{T}$ with $T$ words,
the skip-gram model aims to maximize the objective function:
\begin{equation}\label{eq:skip-gram-obj}
    {\cal Q} 
= \frac{1}{T} \sum_{t=1}^{T}{\sum_{-c \le j \le c, j \ne 0}{\log p(w_{t+j}|w_{t})}}
\end{equation}
where $c$ is the size of context windows, $w_{t}$ denotes the input central word and $w_{t+j}$ means for its neighbouring word. The skip-gram model computes the conditional probability $p(w_{t+j}|w_{t})$ using softmax function:
\begin{equation}\label{eq:word2vec-prob}
    p(w_{t+j}|w_{t}) = \frac{ \exp(\mathbf{w}_{t+j}^{(2)} \cdot \mathbf{w}_{t}^{(1)}) }
    {\sum_{k=1}^{V}{ \exp(\mathbf{w}_{k}^{(2)} \cdot \mathbf{w}_{t}^{(1)}) }}
\end{equation}
where $\mathbf{w}_{t}^{(1)}$ and $\mathbf{w}_{k}^{(2)}$ denote row vectors in matrices ${\bf W}^{(1)}$ and ${\bf W}^{(2)}$ corresponding to word $w_{t}$ and $w_{k}$ respectively.

In this paper, we proposed to incorporate the commonsense knowledge as constraints into the word embedding training process.
Assume the knowledge is represented by a large number of inequalities, denoted as the set $S$.
This paper denotes $s_{ij} = \mbox{sim}(\mathbf{w}_{i}^{(1)}, \mathbf{w}_{j}^{(1)})$ as the semantic similarity hereafter.
The final objective function becomes:
\begin{equation}
	\{ {\bf W}^{(1)}, {\bf W}^{(2)}\} = \arg\max_{{\bf W}^{(1)}, {\bf W}^{(2)}} \;
	{\cal Q} ({\bf W}^{(1)}, {\bf W}^{(2)})
\end{equation}
subject to
\begin{equation}
s_{ij} > s_{kg} \;\;\; \forall ( i,j,k,g )  \in S.
\end{equation}
In this work, we formulate the above constrained optimization problem into an unconstrained one by casting all the constraints as a penalty term in the objective function:
\begin{equation}\label{eq:obj-final}
    \begin{split}
    & {\cal Q}' = {\cal Q} - \beta \cdot {\cal D}\\
    & {\cal D} = \sum_{(i,j,k,g)\in S} f(i,j,k,g)
    \end{split}
\end{equation}
where $\beta$ is a control parameter to balance the contribution of the penalty term in the optimization process.
The function $f(\cdot)$ is a normalization function.
This paper uses a hinge loss function like $f(i,j,k,g) = h(s_{kg} - s_{ij})$ where $h(x) = \max(0, x)$.
Finally, the objective function in eq. (\ref{eq:obj-final}) could be optimized using the standard stochastic gradient descent (SGD) algorithm.
In this paper, we use the open toolkit available on \url{https://github.com/iunderstand/SWE} to implement the KEE models.

\section{PDP Problem Solvers Based on KEE}
As indicated in Figure \ref{fig:kee-focus}, the commonsense knowledge enhanced embeddings are used for feature extraction.
In order to solve PDP problems based on KEE models, this work proposes two problem solvers.
The first solver is a system constructed using \textit{Unsupervised Semantic Similarity Method} (USSM).
The second solver is a system constructed using \textit{Neural Knowledge Activated Method} (NKAM).
\subsection{Unsupervised Semantic Similarity Method}
\begin{figure}[htb]
  \centering
  \includegraphics[width=9cm]{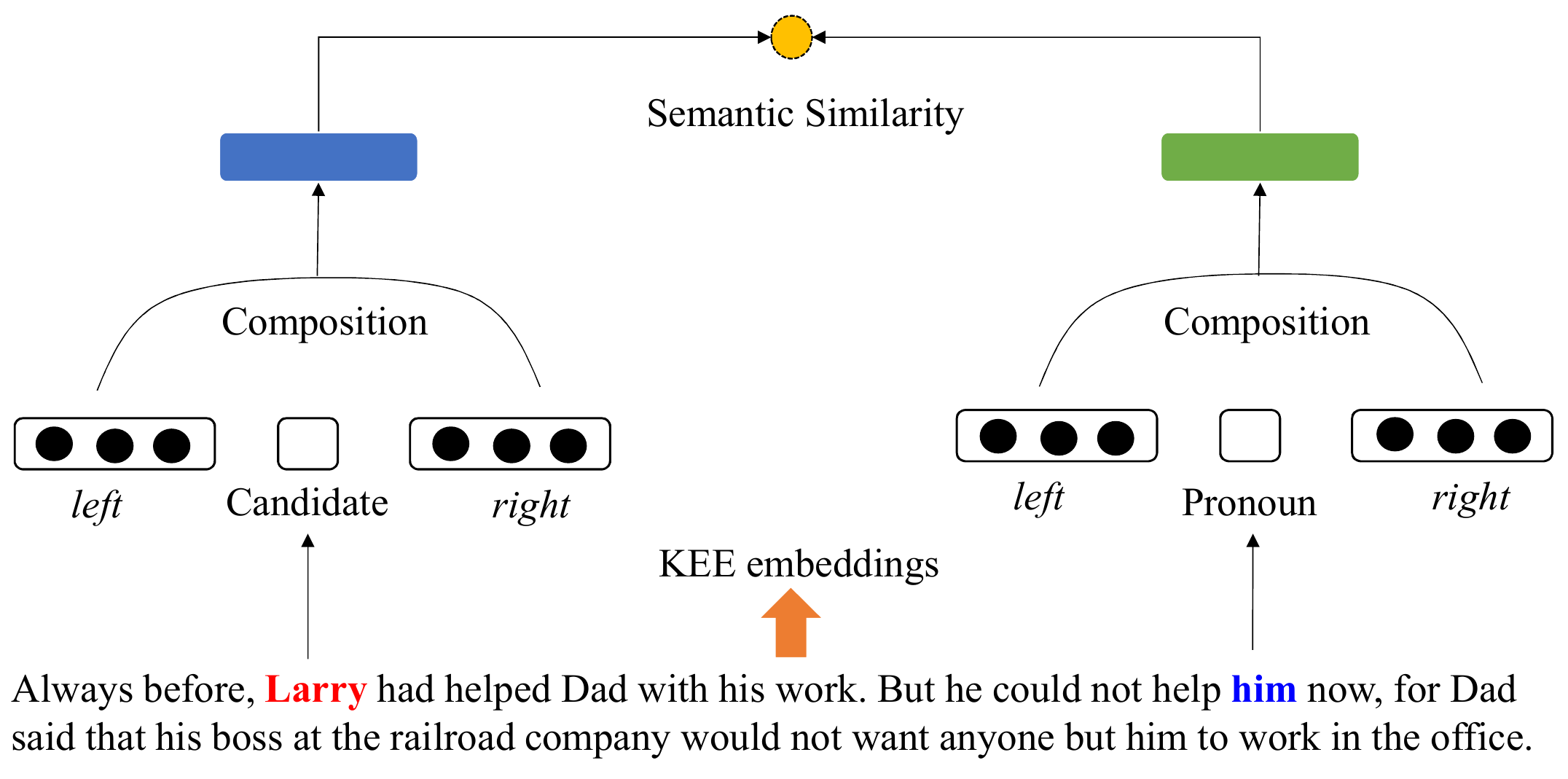}\\ 
  \caption{The unsupervised semantic similarity method proposed to solve the PDP problems.}
  \label{fig:ussm}
\end{figure}

The first method proposed in this paper for answering the PDP problems, shown in Figure \ref{fig:ussm}, is an \textbf{unsupervised} method.
We call it unsupervised semantic similarity method (USSM).
This method aims to represent the pronoun and all the candidate mentions by utilizing the pre-trained KEE embeddings.
For the composition function, we design to use the fixed-size ordinally-forgetting encoding (FOFE) \cite{zhang2015fixed} algorithm, which can almost uniquely encode any variable-length sequence of words into a fixed-size representation.

For one sentence, the function FOFE works as follows.
Given a sequence of words, $S=\{w_{1}, w_{2},..., w_{T}\}$, each word $w_{t}$ is first represented by a 1-of-K representation $\mathbf{e}_{t}$, from the first word $t=1$ to the end of the sequence $t=T$, FOFE encodes each history based on a simple recursive formula (with $\mathbf{z}_{0} = \mathbf{0}$) as:
\begin{equation}\label{eq:fofe}
	\mathbf{z}_{t} = \alpha \cdot \mathbf{z}_{t-1} + \mathbf{e}_{t}, \ (1 \le t \le T)
\end{equation}
where $\mathbf{z}_{t}$ denotes the FOFE code for the partial sequence up to $w_{t}$, and $\alpha, (0 < \alpha < 1)$ is a constant forgetting factor to control the influence of the history on the current position.
Assume we have three words in vocabulary, e.g., $A$, $B$, $C$, whose 1-of-K codes are $[1, 0, 0]$, $[0, 1, 0]$ and $[0, 0, 1]$ respectively.
In this case, the FOFE code for the sequence $\{ABC\}$ is $[\alpha^{2}, \alpha, 1]$, and that of $\{ABCBC\}$ is $[\alpha^{4}, \alpha+\alpha^{3}, 1+\alpha^{2}]$.
In this paper, we first use the FOFE method to encode both left and right contexts of each word into a fixed-size code. Then, we use the embedding matrix $W^{(1)}$, learned by KEE as above, to project them into a low-dimension space. These low-dimensional vectors are used to calculate cosine distances to select the answer from the candidates.

\subsection{Neural Knowledge Activated Method}
\begin{figure}[htb]
  \centering
  \includegraphics[width=8.5cm]{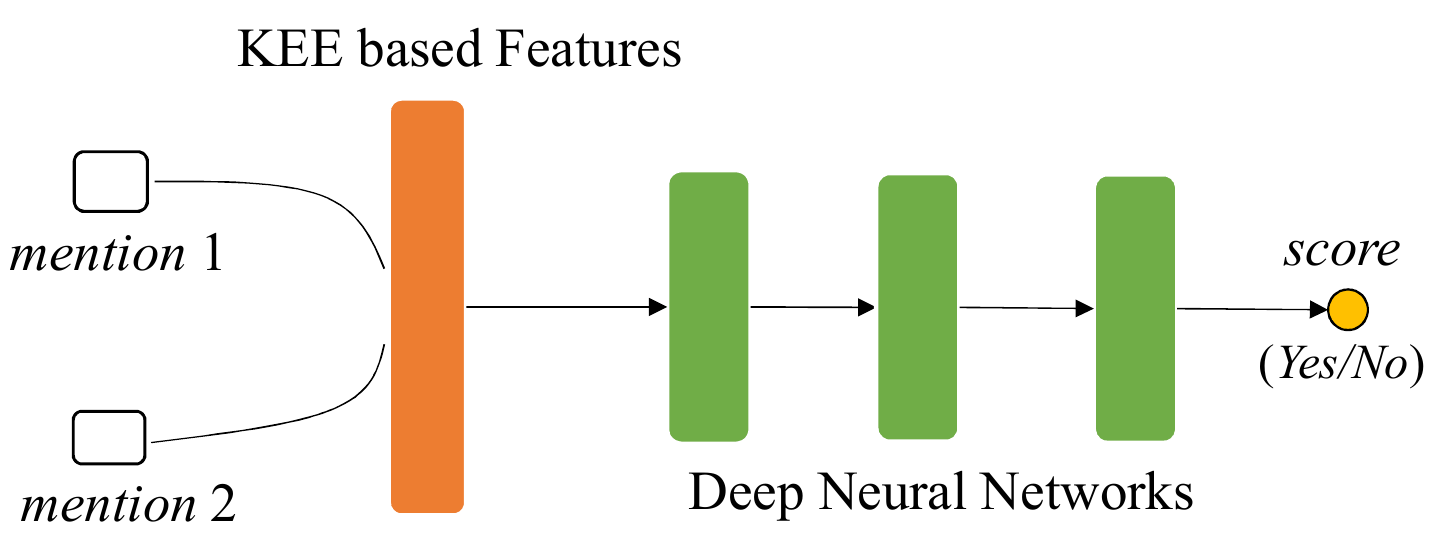}\\ 
  \caption{The neural knowledge activated method proposed to solve the PDP problems.}
  \label{fig:nkam}
\end{figure}

As shown in Figure \ref{fig:nkam}, the second method proposed in this paper is an \textbf{supervised} method.
The difference from the first semantic similarity method is that, it does not simply calculate the semantic similarities between the extracted embedding vectors of the pronoun (to be resolved) and all the candidate mentions, but instead uses the composed embedding vectors as input features and train a deep neural networks (DNN).

The DNN model works as a mention pair classifier for judging whether two mentions are coreferent or not, which is a widely used technology in the coreference resolution community \cite{ng2010supervised}.
Since the features we extracted for training the DNN are composed from the knowledge enhanced embeddings, we call the method as neural knowledge activated method (NKAM) hereafter.

\section{Experiments}
In this section, we present all the experiments conducted to evaluate the effectiveness of the proposed commonsense knowledge enhanced embeddings.
This section starts by introducing the experimental dataset and experimental setup.
After that, experimental results and analysis will be given respectively.
\subsection{Experimental dataset}
The dataset used in this paper is the official test set of the 2016 Winograd Schema Challenge\footnote{Available at \url{www.cs.nyu.edu/faculty/davise/papers/WinogradSchemas/PDPChallenge2016.xml}}.
It contains 60 pronoun disambiguation problems.
The PDP problems are taken from the wild, and from many genres of writing, may \textbf{touch on different aspects of commonsense knowledge} than that which a single person or small group of people could come up with when creating Winograd schemas.
\subsection{Experimental setup}
To make clear all the settings for the proposed methods of this work, we describe the experimental setup as follows.
We will first introduce the setup for knowledge enhanced embeddings, including the training sources and settings.
After that, we will give the descriptions to the setup for PDP problem solvers.
\subsubsection{Setup for knowledge enhanced embeddings}
This paper uses two text corpora to train the knowledge enhanced embeddings. The first one is a book corpus while the second one is a Wikipedia text corpus.
\begin{enumerate}[(1)]
\item \textbf{\textit{CBTest}}. The first corpus is a book corpus collected from the Project Gutenberg\footnote{\url{https://www.gutenberg.org/}}. More specifically, we use the text corpus from the work of \cite{hill2015goldilocks}. After text normalization, the corpus contains 300 million tokens while the corresponding vocabulary contains 53541 words. We call this corpus \textit{CBTest} in this paper.
\item \textbf{\textit{Wikipedia}}. The second corpus is a snapshot of the Wikipedia articles from \cite{shaoul2010westbury}, named as \textit{Wikipedia} in our experiments. The corpus has been pre-processed by removing all the HTML meta-data and hyper-links and replacing the digit numbers with English words. After text normalization, the \textit{Wikipedia} corpus contains about 1 billion tokens, for which we create a lexicon of 235,167 words, each appearing more than 60 times.
\end{enumerate}

Meanwhile, to train the KEE models by incorporating commonsense knowledge, the knowledge bases we used are listed as follows:
\begin{enumerate}[(1)]
\item \textbf{\textit{ConceptNet}}. We use the latest ConceptNet KB for experiments\footnote{\url{http://conceptnet5.media.mit.edu/}}. The concepts defined in ConceptNet contain both word and phrase style, e.g. ``\textit{dream}" and ``\textit{act in play}". In this work, we extract all the commonsense triples defined over all the word level concepts. Based on the extracted triples, we use the rule described in section \ref{sec:rules} to generate knowledge constraints. The number of the generated semantic similarity inequalities is 543,540.
\item \textbf{\textit{WordNet}}. The WordNet we used in this work is WordNet 3.1\footnote{\url{https://wordnet.princeton.edu/wordnet/download/}}. The process to create knowledge constraints would be started by extracting all the semantic relationships, especially the synonymous and antonymous, hypernym and hyponym relationships between all the words. After that, we can generate knowledge constraints by applying the rules presented in section \ref{sec:rules}. The number of the generated semantic similarity inequalities is about 433,428.
\item \textbf{\textit{CauseCom}}. We use the cause-effect pair set collected by the work of \cite{liu2016probabilistic}. It contains hundreds of thousands cause-effect pairs extracted from various text corpora. After applying the rule proposed in this paper to generate knowledge constraints, we get about 786,390 similarity inequalities.
\end{enumerate}

In all the experiments of this paper, the settings for KEE are the same.
The embedding dimension is set to be 100 while the context window size $c$ in eq. (\ref{eq:skip-gram-obj}) is set to be 5.
The combination coefficient $\beta$ in eq. (\ref{eq:obj-final}) is set to be 0.01.
The KEE models are trained by the SGD algorithm.
The initial learning rate is set as 0.025 and the learning rate is decreased linearly during the SGD training process.
\subsubsection{Setup for PDP problem solvers}
As for feature extraction in the USSM or NKAM methods, for both pronouns and candidate mentions, the context we utilize for feature extraction is the entire sentence.
Meanwhile, the weight $\alpha$ in eq. (\ref{eq:fofe}) of the FOFE function is set to 0.7 for context composition.
In the USSM method, we use the popular cosine similarity to evaluate the semantic similarity between any two mentions.
On the other hand, for the NKAM method, this paper uses the popular coreference resolution datasets, i.e., OntoNotes \cite{weischedel2013ontonotes}, to extract labelled mention pairs for model training.
Considering the sentence length in the PDP questions is usually less than 3, in this paper, we extract all the labelled mention pairs for pronouns within three adjacent sentences.
We finally extract 306,903 training mention pairs.
Meanwhile, the corresponding neural network has 1 hidden layer with 300 units.
The non-linear activation function is rectified linear unit (ReLU) \cite{nair2010rectified}.
\subsection{Results achieved on the CBTest corpus}
In addition to experimenting the proposed two methods, i.e., USSM and NKAM, we also construct a system by combing the USSM and NKAM methods.
For each pronoun and its candidate mentions, the system combination procedure is implemented by interpolating the scores calculated by the USSM and NKAM method (the interpolation coefficient is 70\% for NKAM and 30\% for USSM).
Table \ref{tab:allres-1} shows the overall results achieved when using the \textit{CBTest} corpus for KEE training.
From the results, when the KEE models are only trained on texts (no commonsense knowledge combined, which is equal to the skip-gram models), the USSM method and the NKAM method achieve 46.7\% and 50.0\% accuracy on the PDP test set.

The results indicate that, when we train KEE by employing commonsense knowledge bases, we get improved performances.
The best performances are achieved by combining all the three commonsense knowledge bases, including ConceptNet, WordNet and CauseCom.
Our three systems, i.e., USSM, NKAM and USSM+NKAM achieve 55.0\%, 61.7\% and 65.0\% accuracy on the PDP test set.
The 65\% accuracy obtained by the combined system is significant better than the system (53.3\%) constructed solely based on the context (without combining commonsense knowledge during KEE training).
\begin{table*}[htb]
\centering
\begin{tabular}{c|c|l|c|c}
  \toprule
  \multirow{2}{*}{Text Corpus} & \multirow{2}{*}{Problem Solver} & \multicolumn{3}{c}{KEE settings Accuracy} \\
  \cline{3-5}
  & & KEE training sources & Accuracy (\%) & Improvements (\%) \\
  \midrule 
  \midrule 
  \multirow{15}{*}{\textit{CBTest}} & \multirow{5}{*}{USSM} & Context & \underline{46.7} & \\
  & & Context + ConcepNet & 53.3 & +14.1\\
  & & Context + WordNet & 51.7 & +10.7 \\
  & & Context + CauseCom & 53.3 & +14.1\\
  & & Context + All KBs & \bf 55.0 & \bf+17.7\\
  \cline{2-5}
  & \multirow{5}{*}{NKAM} & Context & \underline{50.0} & \\
  & & Context + ConcepNet & 60.0 & +20.0\\
  & & Context + WordNet & 58.3 & +16.6 \\
  & & Context + CauseCom & 60.0 & +20.0\\
  & & Context + All KBs & \bf 61.7 & \bf+23.4\\
  \cline{2-5}
  & \multirow{5}{*}{USSM+NKAM} & Context & \underline{53.3} & \\
  & & Context + ConcepNet & 63.3 & +18.7\\
  & & Context + WordNet & 60.0 & +12.6 \\
  & & Context + CauseCom & 61.7 & +15.7\\
  & & Context + All KBs & \bf 65.0 & \bf+21.9\\
  \bottomrule
\end{tabular}
\caption{Overall experimental results achieved when using the \textit{CBTest} corpus. The random-guess accuracy for this dataset is 45\%.}
\label{tab:allres-1}
\end{table*}
\subsection{Results achieved on the Wikipedia corpus}
Table \ref{tab:allres} shows the overall results obtained when using the \textit{Wikipedia} corpus as the training source for KEE models.
Similar to the results shown in Table \ref{tab:allres-1}, we find the proposed methods achieve consistent improvements over all the baselines.
The three systems, i.e., USSM, NKAM and USSM+NKAM achieve 56.7\%, 63.3\% and 66.7\% accuracy on the PDP test set.
\begin{table*}[htb]
\centering
\begin{tabular}{c|c|l|c|c}
  \toprule
  \multirow{2}{*}{Text Corpus} & \multirow{2}{*}{Problem Solver} & \multicolumn{3}{c}{KEE settings Accuracy} \\
  \cline{3-5}
  & & KEE training sources & Accuracy (\%) & Improvements (\%) \\
  \midrule 
  \midrule 
  \multirow{15}{*}{\textit{Wikipedia}} & \multirow{5}{*}{USSM} & Context & \underline{48.3} & \\
  & & Context + ConcepNet & 55.0 & +13.9\\
  & & Context + WordNet & 53.3 & +10.4 \\
  & & Context + CauseCom & 55.0 & +13.9\\
  & & Context + All KBs & \bf 56.7 & \bf+17.4\\
  \cline{2-5}
  & \multirow{5}{*}{NKAM} & Context & \underline{51.7} & \\
  & & Context + ConcepNet & 60.0 & +16.0\\
  & & Context + WordNet & 60.0 & +16.0 \\
  & & Context + CauseCom & 61.7 & +19.3\\
  & & Context + All KBs & \bf 63.3 & \bf+22.4\\
  \cline{2-5}
  & \multirow{5}{*}{USSM+NKAM} & Context & \underline{53.3} & \\
  & & Context + ConcepNet & 63.3 & +18.7\\
  & & Context + WordNet & 61.7 & +15.7 \\
  & & Context + CauseCom & 65.0 & +21.9\\
  & & Context + All KBs & \bf 66.7 & \bf+25.1\\
  \bottomrule
\end{tabular}
\caption{Overall experimental results achieved when using the \textit{Wikipedia} corpus. The random-guess accuracy for this dataset is 45\%.}
\label{tab:allres}
\end{table*}

\subsection{Final remarks}
The method proposed in this paper have been applied to our system submitted to the 2016 Winograd Schema Challenge. The system we constructed for the competition utilized a small set of CauseCom knowledge. The KEE model was trained on a smaller Wikipedia text corpus. For the problem solver, we selected to use the NKAM method. This system achieved 58\% accuracy during the competition (all results are listed in Table \ref{tab:soa}).
As confirmed by the organizer, 58.3\% was the best performance on this competition\footnote{\url{http://www.cs.nyu.edu/faculty/davise/papers/WinogradSchemas/WS.html}}.
\begin{table}[htb]
\centering
\begin{tabular}{c|c|c}
  \hline
  \multicolumn{2}{c|}{Systems} & Accuracy\\
  \hline
  \multirow{4}{*}{Performance in 2016 Winograd Schame Challenge\footnote{\url{http://www.cs.nyu.edu/faculty/davise/papers/WinogradSchemas/WS.html}}} & Denis Robert & 31.7\%\\
  & Patrick Dhondt & 45.0\% \\
  & Nicos Issak & 48.3\%\\
  & Quan Liu & 58.3\%\\
  \hline
\end{tabular}
\caption{State-of-the-art performances on the PDP test set. All numbers are obtained from the 2016 Winograd Schema Challenge \cite{WSC2016}.}
\label{tab:soa}
\end{table}

Moreover, to the best of our knowledge, the 66.7\% accuracy obtained in this paper is a new state-of-the-art performance on the challenging PDP
task.


%

\section{Conclusions}
This paper proposes commonsense knowledge enhanced embeddings (KEE) toward solving the complex Pronoun Disambiguation Problems (PDP) in Winograd Schema Challenge.
The KEE provides a flexible framework to learn distributed representations under the supervision of commonsense knowledge from large text corpus.
Using the KEE models, we further proposes two methods, i.e. unsupervised semantic similarity method and neural knowledge activated method, to solve the PDP problems.
Experiments conducted on the official dataset show that these KEE models achieve consistent improvements over the baseline systems.
The best performance on the challenging dataset reaches to 66.7\%, which is a new state-of-the-art performance.

\section{Acknowledgements}
This paper was supported in part by the Science and Technology Development of Anhui Province, China (Grants No. 2014z02006), the Fundamental Research Funds for the Central Universities (Grant No. WK2350000001) and the Strategic Priority Research Program of the Chinese Academy of Sciences (Grant No. XDB02070006).

\section*{References}

\bibliography{mybibfile}

\end{document}